\documentclass[conference]{IEEEtran}
\IEEEoverridecommandlockouts
\usepackage{cite}
\usepackage{stfloats}
\usepackage{amsmath,amssymb,amsfonts}
\usepackage{algorithmic}
\usepackage{graphicx}
\usepackage{textcomp}
\usepackage{xcolor,url}

\usepackage{textcomp}
\usepackage{subcaption}
\usepackage{stfloats}
\usepackage{url}
\usepackage{verbatim}
\usepackage{graphicx}
\usepackage{cite}
\usepackage[ruled,vlined]{algorithm2e}

\def\BibTeX{{\rm B\kern-.05em{\sc i\kern-.025em b}\kern-.08em
    T\kern-.1667em\lower.7ex\hbox{E}\kern-.125emX}}
\begin{document}

\title{Interpreting Multi-objective Evolutionary Algorithms via \textit{Sokoban} Level Generation}

\author{\IEEEauthorblockN{Qingquan Zhang\IEEEauthorrefmark{1}, Yuchen Li\IEEEauthorrefmark{1}, Yuhang Lin\IEEEauthorrefmark{1}, Handing Wang\IEEEauthorrefmark{2}, Jialin Liu\IEEEauthorrefmark{1}}
\IEEEauthorblockA{\IEEEauthorrefmark{1}\textit{Guangdong Key Laboratory of Brain-inspired Intelligent Computation, 
Department of Computer Science and Engineering,} \\ \textit{Southern University of Science and Technology, Shenzhen, China}}
\IEEEauthorblockA{\IEEEauthorrefmark{2}\textit{School of Artificial Intelligence, Xidian University, Xi'an, China}
}
\thanks{Q. Zhang and Y. Li contributed equally to this work.}
}

\maketitle

\begin{abstract}
This paper presents an interactive platform to interpret multi-objective evolutionary algorithms. \textit{Sokoban} level generation is selected as a showcase for its widespread use in procedural content generation. By balancing the emptiness and spatial diversity of \textit{Sokoban} levels, we illustrate the improved two-archive algorithm, Two\_Arch2, a well-known multi-objective evolutionary algorithm. Our web-based platform integrates Two\_Arch2 into an interface that visually and interactively demonstrates the evolutionary process in real-time. Designed to bridge theoretical optimisation strategies with practical game generation applications, the interface is also accessible to both researchers and beginners to multi-objective evolutionary algorithms or procedural content generation on a website. Through dynamic visualisations and interactive gameplay demonstrations, this web-based platform also has potential as an educational tool.
\end{abstract}

\begin{IEEEkeywords}
Procedural Content Generation, Multi-objective Optimisation, Multi-objective Evolutionary Algorithms, Two\_Arch2
\end{IEEEkeywords}

\section{Introduction}
\begin{figure}[htbp]
    \centering
    \includegraphics[scale=0.4]{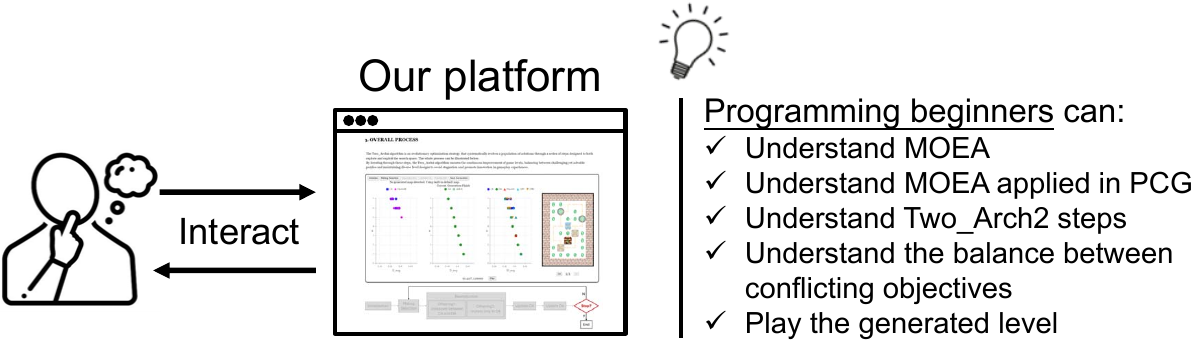}
    \caption{Contributions of our web-based platform}
    \label{fig:contributions}
\end{figure}

Multi-objective optimisation problems (MOPs) are widespread across various domains~\cite{Li20215Many}, such as game optimisation~\cite{volz2019single}, aero-engine calibration~\cite{liu2022fsde_2022}, trauma system design~\cite{wang2016data}, and mitigating bias in decision making~\cite{Mitigating_unfairness_2023}. The literature demonstrates that multi-objective evolutionary algorithms (MOEAs)~\cite{Li20215Many} can effectively address MOPs and produce diverse, high-quality solutions. However, studying MOEAs can be challenging for those unfamiliar with MOEAs. Motivated by this, we aim to develop a web-based interactive platform that visually demonstrates the optimisation process of MOEAs. This platform serves as a bridge between theoretical research and practical applications while meeting the needs of educators, students, researchers, and practitioners.
Procedural content generation (PCG) often involves multi-objective optimisation challenges~\cite{khalifa2020multi,lara2014balance,liu2021deep}, such as balancing desired characteristics in game levels, like emptiness~\cite{Jiang2022Learning} and diversity~\cite{Yannakakis2005Generic,li2024measuring}. Moreover, the well-defined problem and the visually understandable solutions make PCG an ideal testbed for elucidating the mechanisms of MOEAs.
Our web-based platform considers searching for levels of \textit{Sokoban}, a puzzle game in which players push boxes onto targets, with design preferences such as minimalistic (emptiness) and diverse content (spatial diversity).
Our platform integrates Two\_Arch2~\cite{wang2014two_arch2}, a popular and effective MOEA, and interactively demonstrate its optimisation steps. Users can better understand the algorithm and its impact on game content and gameplay by interacting with our platform and through level visualisations, as shown in Fig.~\ref{fig:contributions}.

\section{Concept and Design}
Section \ref{sec:algo} briefly introduces Two\_Arch2 and how it is adapted to \textit{Sokoban} level generations. Section \ref{sec:features} details the design features for enhancing user understanding. The platform is publically accessible at \url{https://aingames.cn/demo/mopcg/index.html}.

\subsection{Searching for diverse game levels with Two\_Arch2}\label{sec:algo}
\begin{figure}[htbp]
    \centering
    \includegraphics[width=0.7\columnwidth]{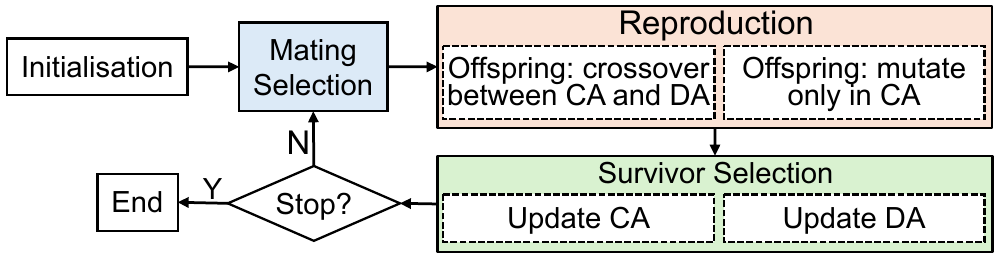}
    \caption{Algorithmic process of Two\_Arch2}
    \label{fig:twoarchive2}
\end{figure}

Our website is designed for visualising and searching for game content through MOEAs, particularly Two\_Arch2~\cite{wang2014two_arch2}. 
As shown in Fig.~\ref{fig:twoarchive2}, Two\_Arch2 adheres to the foundational principles of MOEAs, involving mating selection, reproduction, and survivor selection. The distinctive features of Two\_Arch2 involve maintaining two archives during the evolutionary process: \textit{convergence archive (CA)}, which guides the population toward the \textit{Pareto front (PF)} in terms of convergence, and \textit{diversity archive (DA)}, which ensures a rich diversity within the population, crucial for avoiding local optima.
Besides the playability, we focus on spatial diversity (denoted as $f_{div}$)~\cite{Yannakakis2005Generic} and emptiness (denoted as $f_{emp}$)~\cite{Jiang2022Learning} to address multi-dimensional diversity~\cite{li2024measuring}, formulated as 
\begin{equation}
    maximise \quad \{f_{emp}(level),~f_{div}(level)\},
\end{equation}
where $f_{emp}$ represents the percentage of empty tiles within a \textit{Sokoban} $level$, and $f_{div}$, defined as $-\frac{1}{\log n} \sum_{i=1}^{n}\frac{\alpha p_i}{n}\log \frac{\alpha p_i}{n}$ within $[0,1]$. A $level$ is segmented into $n$ parts, with each row representing a segment. $p_i$ denotes the proportion of empty spaces within each segment, and $\alpha$ is a normalisation factor, ensuring that $\sum_{i=1}^{n}\alpha p_i=1$. A higher $f_{div}$ indicates greater spatial diversity within a level, while a higher $f_{emp}$ suggests the level is more minimalistic.
Fig.~\ref{fig:tradeoffs} is an example captured from our website, a PF is observed in the space of emptiness and spatial diversity through Two\_Arch2 through our website.

\subsection{Website Features}\label{sec:features}
\begin{figure}[htbp]
    \centering
\includegraphics[width=.9\columnwidth]{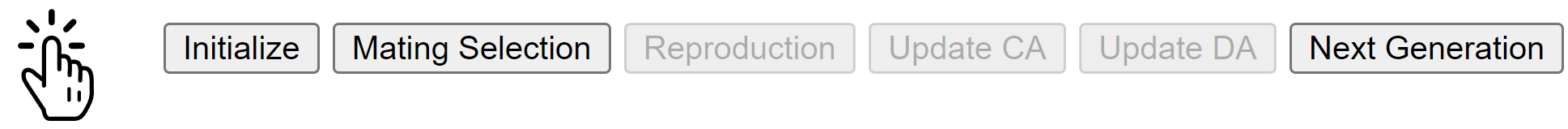}
    \caption{Example of symbol and buttons for interactive content}
    \label{fig:interactive}
\end{figure}

The website integrates the process of generating \textit{Sokoban} levels via Two\_Arch2 into a user-friendly interface that visualises the evolutionary process step by step. All interactive content is highlighted with the same \textit{gesture} as in the example shown in Fig. \ref{fig:interactive}. The design features include the following. 

\subsubsection{Background introduction and game settings} 
The website starts with a basic introduction to MOEA and \textit{Sokoban} levels represented as chromosomes. Users can customise the level parameters, such as size and layouts. An interface is designed for real-time modifications of the chromosome representations in response to the customised changes.

\subsubsection{Dynamic visualisation} 
The current evolutionary process is dynamically highlighted in a chart of the overall process and one or more scatter graphs, characterised by emptiness and spatial diversity, representing the optimised level sets. Each point on a scatter graph represents a level individual, showcasing the progression, and providing visual feedback on the optimisation process, as shown in Fig. \ref{fig:tradeoffs}.
\subsubsection{Step-by-step guidance and overview} 
\begin{figure}[htbp]
\centering
\begin{subfigure}{.3\textwidth}
    \centering
    \includegraphics[width=.7\linewidth]{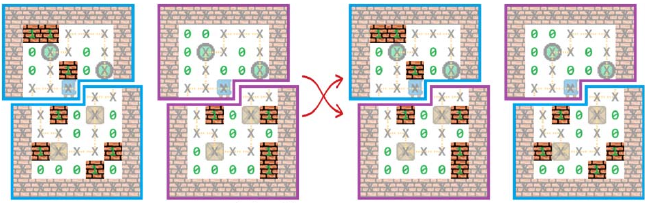}  
    \caption{Crossover}
   \label{fig:crossover}
\end{subfigure}
\begin{subfigure}{.16\textwidth}
    \centering
    \includegraphics[width=.7\linewidth]{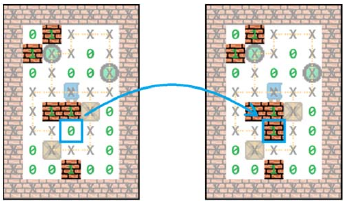}  
    \caption{Mutation}
   \label{fig:mutation}
\end{subfigure}\\
\begin{subfigure}{.2\textwidth}
    \centering
    \includegraphics[width=.8\linewidth]{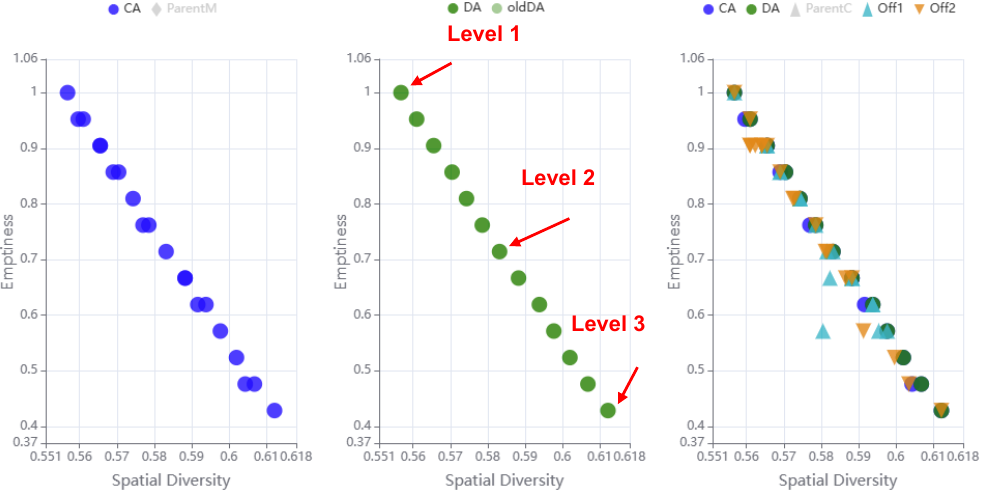}  
    \caption{PF (DA)}
    \label{fig:tradeoffs}
\end{subfigure}
\begin{subfigure}{.2\textwidth}
    \centering
    \includegraphics[width=.8\linewidth]{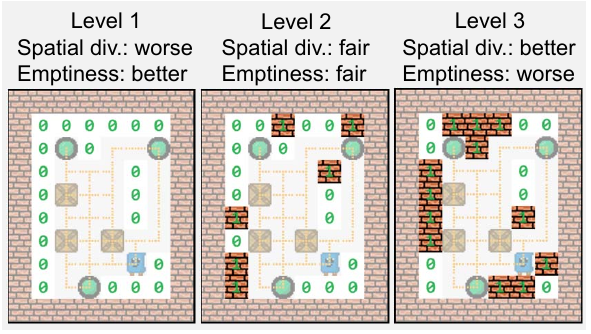}  
    \caption{Levels 1, 2 and 3}
    \label{fig:levels}
\end{subfigure}
\caption{Screenshots of some interactive content}
\label{fig:content}
\end{figure}

The demo provides a step-by-step explanation of Two\_Arch2 and includes information about each component, from chromosome representation, mating selection, reproduction (crossover and mutation, cf. Figs. \ref{fig:crossover} and \ref{fig:mutation}), to survivor selection processes. This educational strategy is designed to simplify complex optimisation processes for those new to evolutionary algorithms, multi-objective optimisation, or procedural level generation. It features tangible visualisations of \textit{Sokoban} levels, enhancing user engagement and understanding. An overview section provides a comprehensive insight into the algorithm and displays the trends in the CA and DA over generations. Users can select any generated level from this overview to engage with in subsequent gameplay sessions designed by~\cite{Zhao2024Playing}, thereby deepening their understanding of the objectives of emptiness and spatial diversity.

\section{Conclusion}

Our platform offers an interactive tool that enables diverse audiences to practically explore various fields, including multi-objective optimisation, evolutionary computation, and search-based level generation, making theoretical outcomes tangible. The platform is also suitable for various educational scenarios such as self-study and classroom use. In the future, it will enhance our platform by implementing various algorithms, allowing for a broader range of learning opportunities.
666t665y
\bibliographystyle{IEEEtran}

\bibliography{refs}

\begin{thebibliography}{10}
\providecommand{\url}[1]{#1}
\csname url@samestyle\endcsname
\providecommand{\newblock}{\relax}
\providecommand{\bibinfo}[2]{#2}
\providecommand{\BIBentrySTDinterwordspacing}{\spaceskip=0pt\relax}
\providecommand{\BIBentryALTinterwordstretchfactor}{4}
\providecommand{\BIBentryALTinterwordspacing}{\spaceskip=\fontdimen2\font plus
\BIBentryALTinterwordstretchfactor\fontdimen3\font minus \fontdimen4\font\relax}
\providecommand{\BIBforeignlanguage}[2]{{%
\expandafter\ifx\csname l@#1\endcsname\relax
\typeout{** WARNING: IEEEtran.bst: No hyphenation pattern has been}%
\typeout{** loaded for the language `#1'. Using the pattern for}%
\typeout{** the default language instead.}%
\else
\language=\csname l@#1\endcsname
\fi
#2}}
\providecommand{\BIBdecl}{\relax}
\BIBdecl

\bibitem{Li20215Many}
B.~Li, J.~Li, K.~Tang, and X.~Yao, ``Many-objective evolutionary algorithms: A survey,'' \emph{ACM Computing Survey}, vol.~48, no.~1, sep 2015.

\bibitem{volz2019single}
V.~Volz, B.~Naujoks, P.~Kerschke, and T.~Tu\v{s}ar, ``Single- and multi-objective game-benchmark for evolutionary algorithms,'' in \emph{Proceedings of the Genetic and Evolutionary Computation Conference}.\hskip 1em plus 0.5em minus 0.4em\relax ACM, 2019, p. 647–655.

\bibitem{liu2022fsde_2022}
J.~Liu, Q.~Zhang, J.~Pei, H.~Tong, X.~Feng, and F.~Wu, ``{fSDE}: efficient evolutionary optimisation for many-objective aero-engine calibration,'' \emph{Complex \& Intelligent Systems}, vol.~8, no.~4, pp. 2731--2747, 2022.

\bibitem{wang2016data}
H.~Wang, Y.~Jin, and J.~O. Jansen, ``Data-driven surrogate-assisted multiobjective evolutionary optimization of a trauma system,'' \emph{IEEE Transactions on Evolutionary Computation}, vol.~20, no.~6, pp. 939--952, 2016.

\bibitem{Mitigating_unfairness_2023}
Q.~Zhang, J.~Liu, Z.~Zhang, J.~Wen, B.~Mao, and X.~Yao, ``Mitigating unfairness via evolutionary multiobjective ensemble learning,'' \emph{IEEE Transactions on Evolutionary Computation}, vol.~27, no.~4, pp. 848--862, 2023.

\bibitem{khalifa2020multi}
A.~Khalifa and J.~Togelius, ``Multi-objective level generator generation with {Marahel},'' in \emph{Proceedings of the 15th International Conference on the Foundations of Digital Games}.\hskip 1em plus 0.5em minus 0.4em\relax ACM, 2020, pp. 1--8.

\bibitem{lara2014balance}
R.~Lara-Cabrera, C.~Cotta, and A.~J. Fern{\'a}ndez-Leiva, ``On balance and dynamism in procedural content generation with self-adaptive evolutionary algorithms,'' \emph{Natural Computing}, vol.~13, pp. 157--168, 2014.

\bibitem{liu2021deep}
J.~Liu, S.~Snodgrass, A.~Khalifa, S.~Risi, G.~N. Yannakakis, and J.~Togelius, ``Deep learning for procedural content generation,'' \emph{Neural Computing and Applications}, vol.~33, no.~1, pp. 19--37, 2021.

\bibitem{Jiang2022Learning}
Z.~Jiang, S.~Earle, M.~Green, and J.~Togelius, ``Learning controllable {3D} level generators,'' in \emph{Proceedings of the 17th {{International Conference}} on the {{Foundations}} of {{Digital Games}}}.\hskip 1em plus 0.5em minus 0.4em\relax ACM, 2022, pp. 1--9.

\bibitem{Yannakakis2005Generic}
G.~N. Yannakakis and J.~Hallam, ``A generic approach for obtaining higher entertainment in predator/prey games,'' \emph{Journal of Game Development}, vol.~1, no.~3, pp. 23--50, 2005.

\bibitem{li2024measuring}
Y.~Li, Z.~Wang, Q.~Zhang, and J.~Liu, ``Measuring diversity of game scenarios,'' \emph{arXiv preprint arXiv:2404.15192}, 2024.

\bibitem{wang2014two_arch2}
H.~Wang, L.~Jiao, and X.~Yao, ``{Two\_Arch2}: An improved two-archive algorithm for many-objective optimization,'' \emph{IEEE Transactions on Evolutionary Computation}, vol.~19, no.~4, pp. 524--541, 2014.

\bibitem{Zhao2024Playing}
Y.~Zhao, C.~Hu, and J.~Liu, ``Playing with {Monte-Carlo} {Tree Search} {[AI-eXplained]},'' \emph{IEEE Computational Intelligence Magazine}, vol.~19, no.~1, pp. 85--86, 2024.

\end{thebibliography}

\end{document}